\documentclass{article}
\usepackage{amsmath,graphicx,mlspconf,booktabs,multirow,xcolor}
\usepackage{xcolor}
\usepackage{comment}

\newcommand{\sa}[1]{{\textcolor{purple} {#1}}}
\newcommand{\omid}[1]{{\color{blue}#1}}

\title{Solving Jigsaw Puzzles in the Wild: Human-Guided Reconstruction of Cultural Heritage Fragments}

\name{%
    Omidreza Safaei$^{1}$%
    \quad Sinem Aslan$^{1,2}$%
    \quad Sebastiano Vascon$^{1}$%
    \quad Luca Palmieri$^{1}$ \\%
    \quad \textit{Marina Khoroshiltseva}$^{1}$%
    \quad \textit{Marcello Pelillo}$^{1,3}$\thanks{This work has received funding from the European Union’s Horizon 2020 research and innovation programme under grant agreement \textit{No 964854}.}%
}

\address{%
    $^{1}$ DAIS, Ca’ Foscari University of Venice, Venice, Italy\\%
     $^{2}$ Department of Historical Studies, University of Milan, Milan, Italy \\
     $^{3}$ Zhejiang Normal University, Jinhua, Zhejiang Province, China%
     }

\name{Omidreza Safaei$^{1}$, Sinem Aslan$^{1,2}$, Sebastiano Vascon$^{1}$, Luca Palmieri$^{1}$, Marina Khoroshiltseva$^{1}$, Marcello Pelillo$^{1,3}$\thanks{This work has received funding from the European Union’s Horizon 2020 research and innovation programme under grant agreement, \textit{No 964854}.}}

\address{$^{1}$ DAIS, Ca' Foscari University of Venice, Venice, Italy \\
$^{2}$ Department of Historical Studies, University of Milan, Milan, Italy \\
$^{3}$ Zhejiang Normal University, Jinhua, Zhejiang Province, China}

\begin{document}

\maketitle

\begin{abstract}
Reassembling real-world archaeological artifacts from fragmented pieces poses significant challenges due to erosion, missing regions, irregular shapes, and large-scale ambiguity. Traditional jigsaw puzzle solvers, ie., often designed for clean, synthetic scenarios, struggle under these conditions, especially as the number of fragments grows into the thousands, as in the RePAIR benchmark. In this paper, we propose a human-in-the-loop (HIL) puzzle-solving framework designed to address the complexity and scale of real-world cultural heritage reconstruction. Our approach integrates an automatic relaxation-labeling solver with interactive human guidance, allowing users to iteratively lock verified placements, correct errors, and guide the system toward semantically and geometrically coherent assemblies. We introduce two complementary interaction strategies—Iterative Anchoring and Continuous Interactive Refinement—which support scalable reconstruction across varying levels of ambiguity and puzzle size. Experiments on various RePAIR groups demonstrate that our hybrid approach substantially outperforms both fully automatic and manual baselines in accuracy and efficiency, offering a practical solution for large-scale, expert-in-the-loop artifact reassembly.
\end{abstract}


\section{Introduction}
\label{sec:intro}
Reassembling fragmented cultural artifacts, such as ancient frescoes, pottery, or mosaics, is a critical yet computationally complex task in archaeology, digital heritage, and museum curation. While the objective may seem straightforward, ie., placing fragments back into their original configuration, the reality is far more challenging. Real-world fragments are often heavily deteriorated, incomplete, or separated by missing regions. Moreover, the scale of these reconstructions introduces significant complexity: archaeological sites frequently yield thousands of fragments, many from distinct objects or surfaces. The recently introduced RePAIR benchmark~\cite{tsesmelis2024repair} exemplifies this scale and complexity, aiming to document over 10,000 fragments from mixed ceiling frescoes. As puzzle size and heterogeneity increase, ambiguity, visual noise, and combinatorial explosion make traditional jigsaw solvers that based on geometric or appearance-based matching ineffective.

Although recent research has produced automatic algorithms for synthetic~\cite{derech2021solving, scarpellini2024diffassemble} or small-scale reassembly tasks, these methods often rely on assumptions such as clean edges, consistent color profiles, or complete fragment sets. Such assumptions rarely hold in real-world archaeological contexts, where fragments are irregular, worn, and contextually ambiguous. Automatic solvers, when applied to large or degraded collections, frequently stall or converge to unstable, locally optimal configurations. Even state-of-the-art systems struggle on real-world data: results on the RePAIR benchmark clearly show that human guidance is essential to producing coherent and accurate reconstructions at scale.

In this paper, we propose a hybrid framework that tightly integrates algorithmic inference with human expertise to enable robust puzzle solving in the wild. At its core lies a game-theoretic solver based on relaxation labeling and replicator dynamics~\cite{khoroshiltseva2024nash}, which estimates fragment placements by optimizing pairwise compatibilities over discrete pose spaces. While this solver has proven effective in synthetic settings, we demonstrate that it performs poorly under real-world conditions unless guided interactively.

Our key innovation is to embed human input directly into the optimization loop through a human-in-the-loop (HIL) design. Users can pause the solver at any time, inspect the current configuration, validate correct placements, and correct misaligned fragments. Verified placements are fixed in the solver state, forming meta-fragments that act as new structural anchors for subsequent iterations. This guidance reshapes the optimization landscape, enabling convergence toward globally coherent solutions even in the presence of noise, ambiguity, or missing data.

We implement this system through a custom interactive graphical interface that supports multiple levels of human input. Users can zoom, pan, and inspect fragments closely. Verified pieces can be locked with a single click, while incorrectly placed fragments can be dragged into the correct position. More complex actions, such as resolving small clusters, are also supported. These interactions are efficient and vary in cognitive cost, enabling both low-effort refinements and expert-level corrections within a fluid user experience. Our framework combines \textit{artificial intelligence} with \textit{human expertise}: a relaxation-labeling solver proposes probabilistic placements, while users iteratively validate, correct, and anchor fragments to guide the optimization process toward stable and coherent assemblies. Our main contributions are threefold: (i) we extend a relaxation-labeling solver by integrating human feedback directly into the optimization loop, allowing dynamic updates of the probabilistic configuration; (ii) we introduce two interaction strategies, namely, \textit{Iterative Anchoring} and \textit{Continuous Interactive Refinement}, that support different levels of user control and computational scalability; and (iii) we demonstrate that our framework outperforms both automatic and manual baselines on the RePAIR benchmark in terms of accuracy and efficiency.

\section{Related Work}
\label{sec:related}
Jigsaw puzzle solving has been studied through both automatic methods and interactive tools, yet most fail under real-world conditions due to noise and poor scalability.

\textbf{Automatic solvers} have mostly targeted synthetic puzzles with clean edges and complete pieces. Derech et al.~\cite{derech2021solving} used shape and color-based cues, while DiffAssemble~\cite{scarpellini2024diffassemble} introduced a diffusion-based pose graph. Though effective in constrained setups, these models degrade in performance when applied to degraded, ambiguous fragments. Among these, Khoroshiltseva et al.~\cite{khoroshiltseva2024nash} introduced a game-theoretic solver for synthetically generated puzzles with irregular shapes, formulating the problem as a relaxation labeling process via replicator dynamics. We adopt this method for its modular architecture and suitability for iterative optimization. However, as we show, its performance drops significantly on large, real-world fragment sets like RePAIR~\cite{tsesmelis2024repair}. Our work addresses this by embedding human validation into the optimization loop.

\textbf{Interactive systems} have focused primarily on manual alignment. Early tools~\cite{brown2012tools, cantoni2020javastylosis} supported fragment inspection and pairing but lacked solver integration. PuzzleFixer~\cite{ye2022puzzlefixer} introduced a VR-based interface, while others~\cite{echavarria2021interactive} emphasized educational use with fully manual reassembly. These systems rely heavily on trial-and-error and offer little algorithmic guidance.

In contrast, we propose a hybrid framework that integrates human input directly into solver dynamics. Through meta-fragment creation and two interaction strategies, our system combines scalability with precision, outperforming prior methods on noisy, large-scale archaeological puzzles.

\section{Interactive Puzzle Assembly Framework}
\label{sec:proposal}
To address the limitations of fully automatic methods in assembling real-world puzzles, 
we propose a hybrid framework that tightly integrates a game-theoretic solver with human-in-the-loop (HIL) interaction.

At its core, the system builds on a relaxation labeling formulation (Section~\ref{sec:solver}) that models puzzle solving as a multiplayer game, where fragments act as players optimizing their placements via replicator dynamics. While effective, this method may converge to suboptimal solutions under weak or noisy signals.

To mitigate this, we introduce an interactive control mechanism (Section~\ref{sec:human_loop}) in which users progressively guide the optimization by validating fragment positions, anchoring reliable substructures, and selectively correcting errors. These interventions are directly injected into the solver’s evolving probability distributions, ensuring that human insight constrains and refines the search space over time.

Our framework supports two modes of interaction tailored to different use cases: \textit{Iterative Anchoring (IA)}, which incrementally grows the solution using local subproblems, and \textit{Continuous Interactive Refinement (CIR)}, which allows full-scene editing and dynamic supervision. Together, these components enable robust, scalable, and interpretable reassembly in challenging real-world settings.

\subsection{Automatic Solver via Relaxation Labeling}
\label{sec:solver}
As the automatic solver, we adopt a game-theoretic framework shown to handle fragments of arbitrary shape, size, and orientation, making it well-suited for real-world scenarios involving damaged or irregular pieces \cite{khoroshiltseva2024nash}. The jigsaw problem is modeled as a non-cooperative multiplayer game, where each fragment acts as a \textit{player} and its feasible placements form its \textit{strategy space}. Compatibility between neighboring pieces defines the payoff, and evolutionary dynamics guide the system toward globally consistent solutions. While we integrate this solver into our Human-in-the-Loop (HIL) framework, the approach remains flexible and can accommodate other solvers with similar capabilities.



\textbf{\textit{Problem Setup and Strategy Space.}} 
Let \( \mathcal{P} = \{1, 2, \dots, n\} \) denote the set of \textit{players}, where each player corresponds to a distinct puzzle piece. Each player \( i \in \mathcal{P} \) chooses a \textit{pure strategy} \( s_i = (x_i, y_i, \theta_i) \) from a discrete \textit{strategy space} \( S_i \), where \( (x_i, y_i) \) specifies the planar coordinates of the piece on the puzzle board, and \( \theta_i \) is its rotation angle. 
The \textit{joint strategy space} for the entire system is defined as the Cartesian product \( S = S_1 \times \cdots \times S_n \), which captures all possible combinations of placements and orientations for all pieces. A particular configuration of the puzzle is represented by a \textit{strategy profile} \( s = (s_1, \dots, s_n) \in S \), which assigns a specific strategy to each player. 




\textbf{\textit{Payoff Structure and Compatibility.}} 
The goal of each player is to choose a strategy that \textit{maximizes its payoff}, which reflects the \textit{compatibility} of its current configuration with those of all other players. These payoffs are modeled as additively separable, in the sense that the total payoff for a player is the sum of pairwise interaction terms with every other player. Formally, for a given strategy profile $s \in S$, the payoff for player $i$ is given by: 
\begin{equation}
\pi_i(s) = \sum_{j=1}^n A_{ij}(s_i, s_j)
\label{eq:payoff_function}
\end{equation}

The partial payoff function \( A_{ij}(s_i, s_j) \) quantifies the compatibility between two fragments (or players) \( i \) and \( j \), given their respective configurations \( s_i \) and \( s_j \). This measure can incorporate various cues such as edge continuity~\cite{khoroshiltseva2024nash}, boundary shape alignment, or learned visual similarity. In this work, we employ a fused compatibility score that integrates three complementary criteria: boundary shape similarity, motif alignment, and edge continuity. Owing to the spatial locality of plausible matches, the resulting payoff matrix is sparse, with non-negligible values concentrated around spatially proximate configurations.

This pairwise structure places the overall game in the well-studied class of polymatrix games~\cite{Jan68}, where each player interacts independently with others through pairwise payoff functions.

\textbf{\textit{Solving via Replicator Dynamics.}} 
In our game-theoretic formulation, a puzzle solution corresponds to a \textit{Nash equilibrium}, ie., a strategy profile $s^* = (s_1^*, \dots, s_n^*)$ where no player can improve its payoff by unilaterally deviating:

\begin{equation}
\pi_i(s_i^*, s_{-i}^*) \geq \pi_i(s_i, s_{-i}^*), \quad \forall s_i \in S_i,\ \forall i \in \mathcal{P}
\end{equation}

Here, $\pi_i(s)$ denotes the total payoff of player $i$ given the full configuration $s$, and $s_{-i}^*$ is the set of strategies chosen by all other players. Such equilibria represent globally consistent placements where no piece (player) benefits from altering its position or orientation. This aligns with the concept of consistent labeling in relaxation labeling theory~\cite{MilZuc91}.

To compute Nash equilibria, we adopt a probabilistic framework where each player maintains a distribution over its strategy space. Let $x_{ih}(t)$ denote the probability of player $i$ playing strategy $h \in S_i$ at iteration $t$. 
These probabilities evolve according to discrete-time \textit{replicator dynamics}:

\begin{equation}
\label{eq:replicator}
x_{ih}(t+1) = x_{ih}(t)\, \frac{\pi_{ih}(x(t))}{\sum_k x_{ik}(t)\, \pi_{ik}(x(t))}
\end{equation}

where $\pi_{ih}(x)$ is the expected payoff of strategy $h$:

\begin{equation}
\pi_{ih}(x) = \sum_j \sum_k x_{jk}\, A_{ij}(h,k)
\end{equation}

These dynamics gradually shift probability mass toward strategies with above-average payoffs. Fixed points of this process correspond to Nash equilibria, and strict equilibria are asymptotically stable~\cite{weibull1997evolutionary}, providing a principled mechanism for convergence in puzzle assembly.

\subsection{Human-in-the-Loop Guided Puzzle Assembly}
\label{sec:human_loop}



Our proposed framework extends a previously published game-theoretic puzzle solver (Section~\ref{sec:solver}), which models fragment placements as a multiplayer game solved via replicator dynamics. While this method effectively leverages local compatibilities, its behavior can be unstable under weak signals or ambiguous fragment boundaries. To address these limitations, we introduce a human-in-the-loop mechanism that enables the user to steer the optimization process by validating correct placements and anchoring verified structures. This interactive loop integrates human input into the solver's internal representation by directly modifying its evolving strategy distributions.

The central principle is to incrementally grow the puzzle around subsets of fragments that are confidently placed, which we call \textit{anchors}. Once an anchor (or verified group of fragments, i.e., a \textit{meta-fragment}) is selected, the solver focuses its optimization locally or globally depending on the interaction strategy, and the user can intervene to refine or lock fragments. This supervision is directly embedded in the probability distributions that govern the solver’s behavior.

\subsubsection{Initialization Phase}
\label{sec:initialization}

The assembly process begins with the selection of a seed fragment that serves as the initial anchor. Inspired by archaeological practice, we prioritize fragments that exhibit strong local context, typically those with distinct geometric features and rich color variation near their boundaries, ie., it is a fact that these characteristics provide informative visual cues about neighboring
fragments.

To formalize this, we compute for each fragment \(f\) a composite score:
\begin{equation}
\mathcal{S}(f) = \alpha \cdot P(f) + (1 - \alpha) \cdot C(f),
\end{equation}
where \(P(f)\) captures structural saliency via perpendicular line detection, and \(C(f)\) captures boundary appearance diversity on the fragment's surface. Specifically, \(P(f)\) is computed by applying the Hough Line Transform to a narrow boundary band on the fragment surface to detect line patterns, and then checking for the presence of perpendicular line pairs, i.e., one line segment lying along the fragment's boundary and another forming a \(90^\circ \pm 5^\circ\)
angle with it. This captures structural continuity across edges, which can provide valuable cues about potential neighboring fragments. If such perpendicular line segments are found, \(P(f) = 1\); otherwise, \(P(f) = 0\). For \(C(f)\), a 2D hue-saturation histogram is constructed over a boundary band in HSV color space. The value of \(C(f)\) is then defined as the ratio of non-zero histogram bins to the total number of bins, providing a measure of color heterogeneity along the fragment’s edge.

The top-\(k\) ranked fragments are shown to the user, who selects the most promising one as the initial anchor \(a^*\), which is fixed at the origin of the global coordinate system with zero rotation:
\begin{equation}
L^{(0)} = \{a^*\}, \quad \pi_{a^*}^{(0)}(x,y,\theta) =
\begin{cases}
1, & \text{if } (x,y,\theta) = (0,0,0),\\
0, & \text{otherwise}.
\end{cases}
\end{equation}

This step provides a stable starting configuration and defines the global coordinate frame for all future placements.

\vspace{-0.3cm}
\subsubsection{Interactive Control over Strategy Distributions}

Building on the initialized anchor, we now allow the user to interactively guide the puzzle assembly process. At the core of our solver is a probability distribution \( \pi_k^{(t)}(x, y, \theta) \) for each player \(k\), representing the confidence of placing that fragment at a specific position and orientation. These distributions evolve over time via relaxation labeling, as described in Section~\ref{sec:solver}.

After each iteration, users can inspect the current configuration and lock fragments they deem correctly placed. This forms the updated locked set:
\begin{equation}
L^{(t+1)} = L^{(t)} \cup V^{(t)},
\end{equation}
where \( V^{(t)} \) is the set of newly verified fragments. The updated probability distributions are then:
\begin{equation}
\pi_k^{(t+1)}(x,y,\theta) =
\begin{cases}
\delta_{(x,y,\theta),(x_k^*,y_k^*,\theta_k^*)}, & \text{if } k \in V^{(t)} \\
\pi_k^{(t)}(x,y,\theta), & \text{if } k \in L^{(t)} \\
\frac{1}{|S_k|}, & \text{otherwise}
\end{cases}
\end{equation}

This mechanism ensures that verified placements are frozen, preventing their update in subsequent iterations, while unverified fragments are re-optimized in light of the updated constraints. These locked fragments—individually or grouped—act as new anchors (meta-fragments) that guide subsequent steps.

Here, \( \delta \) is the Kronecker delta function, assigning full certainty to the user-validated configuration \( (x_k^*, y_k^*, \theta_k^*) \). This effectively locks the corresponding fragment’s state for all subsequent iterations, while the remaining, unverified fragments are re-optimized under the updated constraints. By embedding user-confirmed placements directly into the optimization process, the system ensures that validated fragments remain fixed, providing a stable foundation. These locked fragments, ie., whether individually or as clusters, then serve as new anchors (meta-fragments), guiding the placement of remaining pieces in future steps.

\begin{figure*}[t]
  \centering
\includegraphics[width=\linewidth]{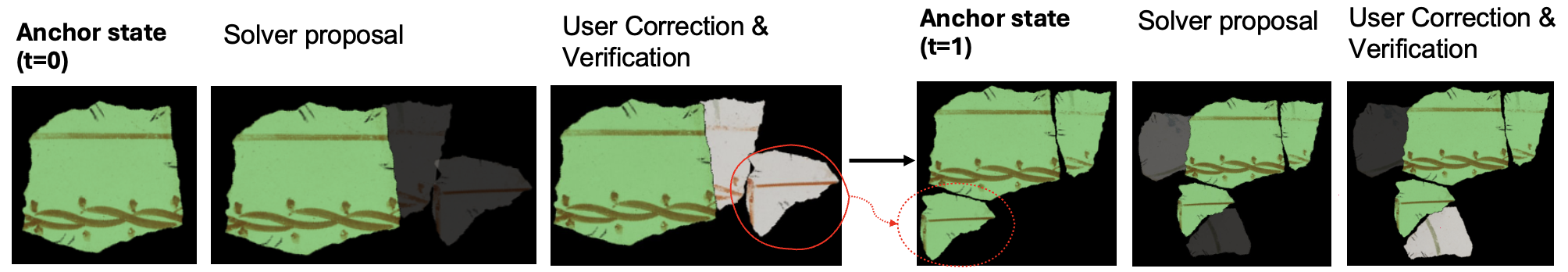}
  \caption{
   Step-by-step visualization of the HIL-IA process. At each iteration \(t\), the current anchor \(L(t)\) is expanded with solver proposals. User verification or correction yields the next anchor \(L(t{+}1)\). Red circles indicate manual adjustments 
   through fragment repositioning that stabilize and guide consistent assembly. 
}
  \label{fig:hil}
\end{figure*}

\subsubsection{Interaction Strategies}
\label{sec:hil-strategies}

The interactive optimization loop described above can be applied in different operational modes, tailored to the scale and ambiguity of the reconstruction task. 

We introduce two complementary strategies, both grounded in the same core mechanism of distribution manipulation. Their primary differences lie in the frequency and scope of user control. Detailed descriptions and comparative analysis of these modes are provided in the following subsections.


\textbf{\textit{Iterative Anchoring (IA).}} This is a localized, scalable approach in which the solver operates over a dynamically defined neighborhood centered around already validated (locked) fragments. Rather than solving globally, the optimization is restricted to a small, top-ranked subset of candidate neighbors, making the process more computationally efficient and user-guided.

In each iteration \(t\), after a set of fragments \(L^{(t)}\) has been locked through user validation, the system identifies candidate neighbors among the remaining fragments \(f \notin L^{(t)}\). For each such fragment, we define its \textit{neighbor suitability score} \(\mathcal{N}(f)\) as the maximum pairwise compatibility it exhibits with any locked fragment:
\[
\mathcal{N}(f) = \max_{j \in L^{(t)}} A_{f j}^{\max},
\]
where
\vspace{-0.3cm}
\[
A_{f j}^{\max} = \max_{(s_f, s_j) \in S_f \times S_j} A_{f j}(s_f, s_j),
\]
and \(A_{f j}(s_f, s_j)\) denotes the compatibility score between fragment \(f\) and fragment \(j\) when placed at discrete poses \(s_f\) and \(s_j\) from their respective strategy spaces. This process effectively scans for fragments that are most likely to adjoin the current assembly, considering all allowable positions and orientations.

Once the neighbor scores \(\mathcal{N}(f)\) are computed, the top-\(k\) highest-scoring fragments form the candidate neighbor set for the next optimization cycle:
\[
N^{(t)} = \left\{ f \mid f \notin L^{(t)}, \text{rank}(\mathcal{N}(f)) \leq k \right\}.
\]
The solver restricts its computation to this subset \(N^{(t)}\), 
optimizing placements relative to the fixed fragments in \(L^{(t)}\). This significantly reduces complexity for the automatic solver and directs human attention toward the most plausible connections.

Although the system automates candidate ranking, the expert retains full control. If a proposed neighbor is rejected, the system automatically advances to the next candidate in the list, maintaining a fluid and effective interaction loop. This iterative expansion process enables scalable puzzle solving and is especially suited for large-scale or high-density scenarios.

\textbf{\textit{Continuous Interactive Refinement (CIR).}}
In contrast to IA, this strategy maintains a global scope. The solver continuously optimizes over all fragments, allowing the user to pause the process, inspect intermediate results, correct any incorrect placements, and resume the optimization. Corrections are enforced by locking the validated configuration, thereby excluding them from further adjustment. This approach is ideal for mid-sized puzzles or regions with high ambiguity, where broader context and global consistency are crucial.

\begin{figure*}[t]
  \centering
  \includegraphics[width=0.9\linewidth]{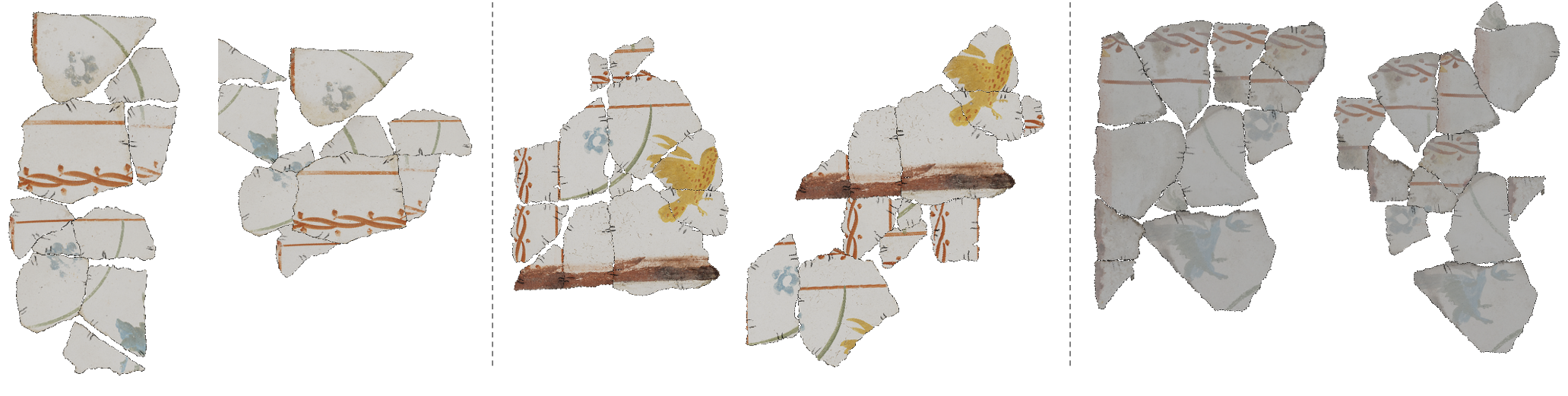}
  \caption{
    Side-by-side visual comparisons between our Human-in-the-Loop solution and the automatic solver \cite{khoroshiltseva2024nash} output across three groups from the RePAIR benchmark \cite{tsesmelis2024repair}. Left: Group 1, Center: Group 3, Right: Group 39. In each pair, the left cluster shows the HIL-IA guided reconstruction, while the right cluster shows the fully automatic result.
  }
  \label{fig:dataset}
\end{figure*}

\vspace{-0.5cm}
\section{Experiments}
\label{sec:experiments}

We evaluate the effectiveness of our Human-in-the-Loop (HIL) puzzle-solving framework on real-world archaeological data. The system is fully implemented in Python using the Kivy library, which provides a responsive and extensible GUI to facilitate expert interaction. Its modular design also enables future extensions, including the integration of inpainting tools or adaptive ranking strategies based on reinforcement learning.

\textbf{\textit{Dataset.}} We conduct experiments on three challenging fresco fragment groups from the RePAIR benchmark~\cite{tsesmelis2024repair}: Groups 1, 3, and 39 (visualized in Figure~\ref{fig:dataset}). These groups are characterized by irregular fragment shapes, eroded edges, and visually complex motifs, capturing the realistic difficulties of large-scale, unconstrained archaeological reconstruction.

\textbf{\textit{Evaluation Metrics.}}
To assess reconstruction accuracy, we adopt the geometry-based $Q_{\text{pos}}$ metric introduced in~\cite{tsesmelis2024repair}. This score measures the normalized overlap between predicted and ground-truth fragment positions, while remaining invariant to global transformations. Additionally, we compute pixel-level Root Mean Square Error ($\text{RMSE}_{\text{px}}$) to quantify Euclidean misalignment. Higher $Q_{\text{pos}}$ and lower $\text{RMSE}_{\text{px}}$ indicate better alignment, and larger fragments are weighted more heavily to reflect their greater significance in the overall structure.



\textbf{\textit{Comparison Methods.}} We evaluate four reconstruction strategies: (i) \textit{Auto RL}, a fully automatic baseline using relaxation labeling~\cite{khoroshiltseva2024nash}; (ii) \textit{HIL-IA}, Iterative Anchoring, where users guide reconstruction by locking verified placements at each local stage; (iii) \textit{HIL-CIR}, Continuous Interactive Refinement, in which the solver runs globally while users pause and adjust the configuration as needed; and (iv) \textit{Manual}, a fully manual reconstruction using the GUI without solver support.

\subsection{Performance evaluation}
\vspace{-0.3cm}
\begin{table}[t]
\centering
\caption{Accuracy and run–time on RePAIR groups.}
\small            
\setlength{\tabcolsep}{3pt}
\renewcommand{\arraystretch}{0.97}
\begin{tabular*}{\columnwidth}{@{\extracolsep{\fill}}l l r r r@{}}
\toprule
\textbf{Grp} & \textbf{Solver} & \textbf{Time} & \textbf{Q-Pos}\,$\uparrow$ & \textbf{RMSE}\,$\downarrow$ \\
& & (s) & & (px) \\ \midrule
\multirow{4}{*}{G1}
  & Auto RL         &  87 & 0.311 & 16.9 \\

  & HIL-IA          & 355 & 0.895 & 0.58 \\
  & HIL-CIR          & 336 &  0.909 & 0.52 \\
  & Manual           & 500 & N/A   & N/A \\[1pt]
  \midrule
\multirow{4}{*}{G3}
  & Auto RL         & 132 & 0.315 & 18.3 \\
  & HIL-IA          & 343 & 0.882& 0.89 \\
  & HIL-CIR          & 290 & 0.877 & 0.71 \\
  & Manual           & 350 & N/A   & N/A \\[1pt]
  \midrule
\multirow{4}{*}{G39}
  & Auto RL         & 120 & 0.197 & 15.6 \\
  & HIL-IA          & 451 & 0.886 & 1.61 \\
  & HIL-CIR         & 267& 0.906& 0.68 \\
  & Manual           & 460 & N/A   & N/A \\
\bottomrule
\end{tabular*}
\label{tab:results}
\end{table}


\textbf{\textit{Quantitative analysis.}} Table~\ref{tab:results} shows that both human-in-the-loop strategies (HIL-IA and HIL-CIR) successfully complete the puzzle, outperforming the automatic relaxation-labeling (Auto RL) solver, which often fails to produce stable assemblies. All $Q_{\text{pos}}$ scores remain below 1, as expected in real-world cases with erosion and ambiguous boundaries. Among HIL methods, HIL-CIR consistently achieves the highest accuracy in both $Q_{\text{pos}}$ and $\text{RMSE}_{\text{px}}$, benefiting from continuous user refinement of the global structure. HIL-IA, in contrast, iteratively anchors local neighborhoods, which improves scalability and solver efficiency but may limit global corrections unless users revisit earlier placements. While individual HIL-IA loops are faster due to their local scope, the total reconstruction time may grow with the number of iterations. Nonetheless, its meta-fragment expansion strategy makes HIL-IA essential for large-scale problems. Both HIL modes offer a practical trade-off between manual effort and solver assistance, and they are substantially more accurate and efficient than both Auto RL and manual-only reconstructions.

\textbf{\textit{Qualitative Analysis.}} 
Figure~\ref{fig:dataset} visually compares our Human-in-the-Loop (HIL) reconstructions to those produced by the automatic relaxation-labeling (RL) solver across three challenging groups from the RePAIR benchmark. In each group, the fully automatic solver produces incomplete solutions, the overall configuration is typically fragmented and lacks global consistency. By contrast, the HIL results (left cluster in each pair) are visually coherent, demonstrating accurate placement and improved motif alignment, even under erosion or missing regions.

The stepwise benefits of our HIL-IA approach are illustrated in Figure~\ref{fig:hil}. At each iteration, the solver proposes new placements for the current fragments. The user then verifies these suggestions, either accepting fragments that are well-placed or manually correcting errors or rejecting them (by not interacting with them). This incremental loop of solver proposal and user verification, stabilizes the assembly process. Even minimal, non-expert feedback suffices to guide the RL solver away from unstable solutions and towards a globally reconstruction.

Overall, these visualizations demonstrate the central advantage of our HIL framework: sparse human intervention is enough to overcome the limitations of purely automatic methods. The result is robust and efficient outperforming both fully automatic and manual baselines, especially in real world puzzles.

\vspace{-0.3cm}
\subsection{Conclusion}

We presented a human-in-the-loop framework for assembling fragmented cultural heritage artifacts, integrating expert guidance with a relaxation labeling solver. Experiments on the RePAIR benchmark showed that our hybrid approach significantly improves reconstruction accuracy and efficiency compared to fully automatic and manual baselines. This work demonstrates the value of interactive optimization in tackling wild real-world puzzles.

\vspace{-0.3cm}
\bibliographystyle{IEEEbib}
\bibliography{strings,refs}

\end{document}